\documentclass[letterpaper, 10 pt, conference]{ieeeconf}
\IEEEoverridecommandlockouts
\usepackage{cite}
\usepackage{amsmath,amssymb,amsfonts}
\usepackage{algorithmic}
\usepackage{graphicx}
\usepackage{textcomp}
\usepackage{xcolor}
\usepackage{hyperref}
\usepackage{float}
\usepackage{svg}
\usepackage{url}
\usepackage{orcidlink}

\title{\LARGE{\bf{Time-Constrained Intelligent Adversaries for \\Automation Vulnerability Testing: A Multi-Robot Patrol Case Study}}}

\author{James C. Ward, Alex Bott, Connor York, and Edmund R. Hunt%
\thanks{This work was supported by the Engineering and Physical Sciences Research Council via the FARSCOPE-TU Centre for Doctoral Training (grant no. EP/S021795/1) and by the Royal Academy of Engineering under the Research Fellowship program.}%
\thanks{All authors are with the School of Engineering Mathematics and Technology, University of Bristol. Email: \{\texttt{james.c.ward}, \texttt{edmund.hunt}\}\texttt{@bristol.ac.uk}}%
}

\begin{document}
\maketitle

\begin{abstract}

Simulating hostile attacks of physical autonomous systems can be a useful tool to examine their robustness to attack and inform vulnerability-aware design. In this work, we examine this through the lens of multi-robot patrol, by presenting a machine learning-based adversary model that observes robot patrol behavior in order to attempt to gain undetected access to a secure environment within a limited time duration. Such a model allows for evaluation of a patrol system against a realistic potential adversary, offering insight into future patrol strategy design. We show that our new model outperforms existing baselines, thus providing a more stringent test, and examine its performance against multiple leading decentralized multi-robot patrol strategies.

\end{abstract}

\section{Introduction}

Security in automated and robotic systems is of increasing importance as these systems becomes more pervasive and integrated throughout society. Beyond the obvious considerations of cybersecurity and communication security, an important facet of this is physical security --- the robustness of these systems to interference in the real world from a hostile actor. The concept of red-teaming in cybersecurity refers to the simulation of a hostile attack in order to evaluate a security system. Typically, this is performed by human experts, but in some cases a sufficiently capable model of an intelligent attacker may be a practical alternative for examining a system's capabilities. In this paper we present a framework for applying this concept to examine the vulnerabilities of a physical automated system --- specifically, through the lens of multi-robot patrol.

Physical security systems can have potential vulnerability to an attacker who, by observing the nature and physical behavior of the system, can learn where and when to attempt to bypass it in order to exploit discovered vulnerabilities. In this paper, we simulate this scenario using our Time-Constrained Machine Learning (TCML) adversary model, which, when presented with an unseen scenario, must learn from scratch in order to develop a model of where and when to attack in order to maximize its probability of success within a fixed time horizon.

\begin{figure}[]
   \centering
   \includegraphics[width=0.87\columnwidth]{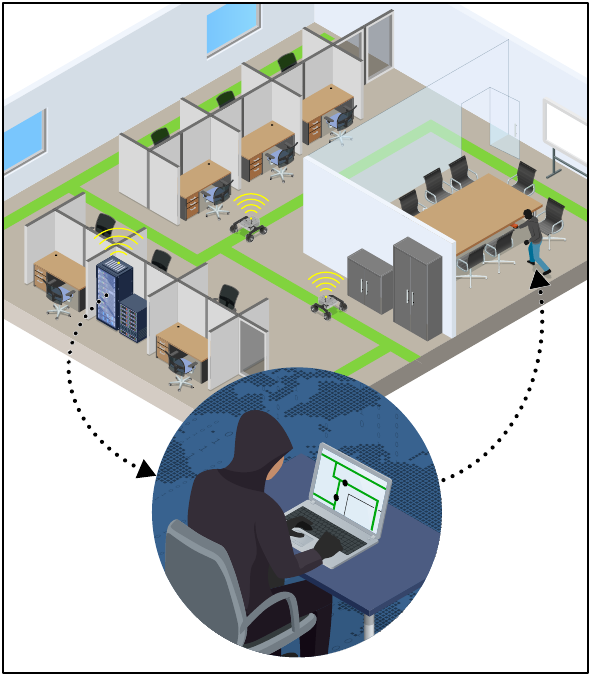}
   \caption{Illustration of an intelligent adversary attacking a patrol system in an indoor security scenario (planting an eavesdropping device)}
   \label{fig:adversary_illustration}
   \vspace{-6mm}
\end{figure}

We apply our TCML adversary to the case of multi-robot patrolling, to demonstrate how it can reveal vulnerabilities in automated patrol systems to better inform strategy design. Such a model could represent a plausible penetration testing simulation in a range of contexts --- any security scenario in which an adversary would have a limited time to make observations before attempting a single attack (as shown in Figure~\ref{fig:adversary_illustration}, for example) could benefit from a similar approach.

An important consideration of the deployment of robotic security systems is the examination of the robots themselves through the lens of cybersecurity~\cite{review_added:1, review_added:2}. While we do not examine this problem in this work, the ability to detect hostile intrusion or unintended behavior in robot teams, especially in security-critical or human interaction scenarios, is essential to the real-world safety of such systems.

\section{Background and related work}

\subsection{Penetration testing robotic systems}

The concepts of penetration testing and red-teaming are not new within the field of robotics, with applications such as stress-testing language-conditioned robotic models~\cite{karnik:foundation_models, abhangi:redteam} and discovering vulnerabilities in learned visuomotor policies~\cite{majumdar:redteam}. However, while valuable, these approaches have little relevance to the traditional idea of red-teaming for security systems. Literature examination of these topics is typically extremely application-specific, and carrying out real-world simulated attacks on a security system is a costly and complex process~\cite{mansfield:redteam}. As such, being able to examine aspects of a physical security system for vulnerabilities in simulation has significant potential value. We propose that robotic security systems are a natural target for simulated red-teaming, as predictable policies present a potential vulnerability that could be exploited by hostile agents. As autonomous robotic systems are inherently more straightforward to simulate than human-centric systems, the smaller reality gap means that conclusions drawn from an intelligent attacker model in simulation can be highly applicable to real physical systems. We examine this further in the context of autonomous robot security by multi-robot patrol.

\subsection{The Multi-Robot Patrolling problem} Here we discuss the background of the Multi-Robot Patrolling (MRP) problem in literature. This problem involves the persistent monitoring of an environment by a team of autonomous mobile robots. This is typically achieved by abstracting the environment as a patrol graph~\cite{machade:idleness_paper}, allowing performance to be measured by considering visitation of the vertices of the graph. The typical key performance metric is ``idleness"~\cite{machade:idleness_paper}, which is measured for a vertex of the patrol graph as the time since the last visit by an agent, giving minimization targets of mean or maximum vertex idleness over an extended period of time. For a single agent, determining an optimal patrol tour for idleness minimization is equivalent to solving the traveling salesman problem on the patrol graph~\cite{chevaleyre:tsp_paper}. For multiple agents, this can be extended to either an approach where multiple agents are evenly spaced around a single tour of the entire graph or a partition-based strategy, whereby a central controller partitions the patrol graph into disjoint subgraphs and then assigns one or more agents to each subgraph. Approximate algorithms exist to generate such partitions~\cite{Afshani2020, Afshani2022}, and a number of well-established algorithms generate approximate solutions to the traveling salesman problem~\cite{Rosenkrantz:tsp}. These approaches allow a central controller to calculate and then allocate tours on the patrol graph to each agent, guaranteeing idleness minimization performance within known bounds of optimality.

However, while centralized solutions may be able to achieve approximately optimal performance, they may be time-consuming and computationally expensive to calculate and cope poorly with agent failure or dynamic environments, as any change to the patrol system would require re-calculating optimal paths to maintain guarantees of performance. These problems can be addressed by decentralized strategies, in which each agent acts based on its own observations and communicates with other agents where possible. Decentralized strategies can offer rapid deployment with no pre-calculation of patrol routes, robustness to agent failure or changing environments, and suitability for communication-limited environments. As such, the design of decentralized strategies for MRP has seen considerable attention --- high-performing approaches include DTAG and DTAP~\cite{farinelli:dta_paper}, both based on dynamic task allocation, the family of Bayesian strategies comprising GBS, SEBS~\cite{portugal:sebs_paper}, and CBLS~\cite{portugal:cbls_paper}, and ER~\cite{yan:er_paper}, a strategy based on predicting vertex idlenesses.

In recent years, learning-based strategies --- especially approaches utilizing graph neural networks (GNNs) --- have become increasingly popular, due to the ability of GNNs to allow learning of inter-agent interactions in highly scalable systems~\cite{li:gnn, zhou:gnn, tolstaya:gnn}. Learning-based controllers for coverage of a graph-structured environment~\cite{tolstaya:gnn_patrol, ward:suns_mns} and variants of MRP~\cite{guo:marl_approach, zhou:learning_patrol} exist, trained with reinforcement learning or imitation learning of expert controllers. Online learning to improve performance in real time has also been presented in decentralized~\cite{portugal:cbls_paper} and modern learning-based~\cite{zhou:learning_patrol} patrol strategies. The ability of these approaches to adapt to dynamic scenarios in real-time varies --- some produce robust decentralized policies that can transfer well to unseen or dynamic environments and team sizes~\cite{portugal:cbls_paper, tolstaya:gnn_patrol, ward:suns_mns}, while some produce policies that are environment-specific or centralized for a fixed number of agents~\cite{guo:marl_approach, zhou:learning_patrol}.

Extensions of this problem in real-world contexts include examinations of efficient localization~\cite{Hsu2014} and energy management~\cite{Ghosh2021} within a patrol scenario, but in this work we do not consider these kinds of extensions due to the level of scenario-specificity required when considering them.

\subsection{Adversarial patrolling}

Here we are concerned with the ability of an automated system (specifically a patrol team) to detect an attacker trying to gain undetected access to the patrol environment, a problem known as ``adversarial" patrolling. Evaluating this ability goes beyond the typical idleness minimization measure of MRP strategy performance.

\subsubsection{Defender models}

As a fully deterministic (and therefore potentially predictable) patrol strategy is at an obvious disadvantage against an intelligent attacker~\cite{agmon:full_knowledge}, many approaches to adversarial patrolling involve generating non-deterministic strategies. These typically appear as Markov chains on the patrol graph~\cite{duan:markov, alam:markov, alam:markov2, basiloco:markov} and are often generated using game theoretical approaches~\cite{yang:gametheory, asghar:markov2}, where the attacker/defender problem can be formulated as a Bayesian Stackelberg game~\cite{parachuri:stackelberg_games, clempner:game_theory}. However, such approaches are almost universally only applicable to a single defender (one exception being~\cite{zhou:learning_patrol}), and require significant precomputation of Markov chains that would be invalidated by any change to the environment. Some non-Markov chain based learning approaches have also attempted to partially optimize for unpredictability~\cite{guo:marl_approach}, but also suffer from learning controllers highly specific to the environment and patrol team size. Constraining the patrol graph to an open or closed polyline, as in the field of fence patrolling, allows adversarial patrolling to be tackled directly for multiple patrol agents~\cite{Buermann:fence_patrol, agmon:fence_patrol, sless:coordinated_attacks, sless:sequential_attacks}, but at the cost of strict environment constraints. 

\subsubsection{Attacker models}

For a defender (patrol system) in an adversarial scenario to be evaluated, it must be tested by an attacker. Various game theoretical attackers are considered in single attacker/single defender scenarios, and some models to attack multi-agent patrol on general graphs have been previously presented, but very few ``intelligent" attacker models have been proposed. Existing non scenario-specific attacker models include \textit{random} intruders~\cite{asghar:markov}, which attack the patrol graph at a random time and place, \textit{deterministic} intruders~\cite{asghar:markov}, which attack a vertex immediately after a patrol agent departs, and \textit{full-knowledge}~\cite{agmon:full_knowledge} intruders, which have knowledge of the algorithms being followed by patrol agents. Finally, following~\cite{asghar:markov}, we define \textit{intelligent} intruders as adversary models that attack the patrol graph at the time and place chosen to maximize expected reward. Previously we developed a learning-based \textit{probabilistic} intelligent intruder in ~\cite{ward:benchmarking_method}. The goal of all attacker models is to spend a set amount of time at a vertex in the patrol graph without visitation by a patrol agent. Here we use random, deterministic, full-knowledge and intelligent (probabilistic) intruders as comparison baselines for our TCML model.

\section{Problem definition and adversary models}
\label{sec:adversary_models}
In keeping with the literature on decentralized MRP, we model our environment as an undirected weighted patrol graph $\mathcal{G}$, where the vertices $\mathcal{V}$ represent points of interest to be repeatedly visited and the edges $\mathcal{E}$ represent traversable routes between points of interest, with weights corresponding to the travel time. The patrol team consists of a set of mobile patrol agents on $\mathcal{G}$ seeking to minimize idleness (either mean or maximum, depending on the patrol strategy) over $\mathcal{V}$ over an extended time horizon, where the instantaneous idleness $I$ of a vertex at time $t$ is defined as the time since it was last visited by a patrol agent. The specific behavior of each patrol agent is defined by the patrol strategy being used. 

To attack this system, we define several models of adversary. All adversary models share the same goal --- to launch a successful attack on a vertex of $\mathcal{G}$ within a time horizon $T$. The attack succeeds if, starting from the time at which the attack is launched, no patrol agent visits the target vertex for a fixed attack duration $\tau$. All adversaries are able to observe the environment and the patrol team in real time, with knowledge of the positions of all patrol agents and idlenesses of all vertices of $\mathcal{G}$ at every timestep since the start of the scenario. We assume the adversaries have the ability to enter the patrol graph at any point without prior detection, to account for the plausible worst-case scenario of a trusted but compromised actor attempting to gain access to a controlled area --- the presence of said actor in the environment may not raise concern until they began their attack.

The adversary models used in this work are as follows:
\begin{enumerate}
    \item \textit{Random} adversary: the adversary will attack a random vertex of $\mathcal{G}$ at a random time $0\leq t \leq T-\tau$.
    \item \textit{Deterministic} adversary: the adversary will attack the first vertex that a patrol agent is observed leaving, as soon as the agent leaves.
    \item \textit{Full-knowledge} adversary: the adversary has full knowledge of the state of the environment and the patrol strategy being used. Therefore,  against deterministic strategies it is able to perfectly predict whether an attack will succeed against a given vertex given the state of the environment at any $t$. In this work we extend this ability to nondeterministic strategies as well, which would not be possible in reality but is useful as a theoretical baseline. As such, the adversary will attack the first vertex against which an attack would succeed, at the time that the attack would succeed. If no attack would succeed against any vertex for any $0 \leq t \leq T$, no attack will be launched.
    \item \textit{Intelligent (Probabilistic)} adversary: using the method presented in~\cite{ward:benchmarking_method}, the adversary observes the environment in order to train a model that calculates an expected probability of an attack succeeding against a given vertex as a function the state of the environment and an expected probability of a given state occurring. From this, it selects a set of ``attack states", i.e. areas of the state-space of the environment in which it will launch an attack against a given vertex in order to maximize its probability of launching a successful attack within the time horizon. To allow this model to function in time-constrained scenarios, we have modified it to utilize the same ``arming" method as used in our new adversary model, described in Section~\ref{sec:method}.
    \item \textit{Intelligent (TCML)} adversary: our time constrained machine learning adversary, see Section~\ref{sec:method}.
    
\end{enumerate}

\section{Time-constrained machine learning adversary}
\label{sec:method}

\begin{figure}[H]
    \centering
    \vspace{-2mm}
    \includegraphics[width=0.75\columnwidth]{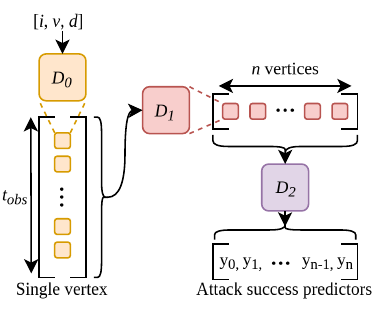}
    \vspace{-4mm}
    \caption{Neural network architecture overview}
    \label{fig:nn}
    \vspace{-4mm}
\end{figure}

The core of our adversary model\footnote{Our implementation of the full TCML adversary model can be found at \url{https://github.com/jward0/Intelligent-Intruder}.} is a neural network, trained on-line from scratch within each attack scenario, that predicts whether an attack against a given vertex of $\mathcal{G}$ will succeed given the state of the environment and patrol team. An overview of this network is shown in Figure~\ref{fig:nn}. To clarify notation --- we refer to the time duration of observations input to the neural network as $t_{obs}$, the current timestep at an arbitrary point in the scenario as $t_{elapsed}$ or $t_{el}$, and as previously stated, we refer to adversary attack duration as $\tau$ and the duration of the time horizon as $T$. 

For a given timestep, the adversary observes the locations of all patrol agents and the idlenesses of $\mathcal{V}$. From these, it calculates a ``distance metric" $d$ for each vertex, defined as the sum of the reciprocals of the distances of all patrol agents to the vertex (calculated along shortest paths through the environment), and a ``velocity metric" $v$ for each vertex, defined as the sum of the velocities towards the vertex divided by distances to the vertex for each patrol agent (where instantaneous velocity is approximated linearly from timestep to timestep, and as before shortest environment paths are used). These metrics and the instantaneous idlenesses $i$ of each vertex for the last $t_{obs}$ timesteps are input to the network. The values associated with each vertex at each timestep are then passed into a 2-layer dense network $D_0$ with output size of 1, with leakyReLU activation on each layer. For each vertex, the resultant values are then passed into a dense layer $D_1$ of output size 1 with leakyReLU activation. Concatenating across all vertices, this results in a vector with size equal to the number of vertices $n$, which is passed into a dense layer $D_2$ with output size of $n$ and sigmoid activation. The weights of this layer have $l_1$ regularization applied, with a regularization factor equal to a set hyperparameter divided by the number of weights in the layer. The values of the output are used as predictions of attack success against each vertex at the given timestep. Hyperparameter values selected are shown in Table~\ref{table:hyperparams}.

\begin{table}[]
    \centering
    \caption{Neural network hyperparameters}
    \vspace{-2mm}
    \begin{tabular}{ll}
        \hline
        Hyperparameter & Value \\ \hline
        Optimizer & Adam \\ 
        Learning rate & 0.001 \\
        ReLU negative slope & 0.3 \\
        Minibatch size & 4 \\ 
        $D_0$ hidden layer size & 6 \\ 
        $D_2$ $l_1$ regularization factor & 0.1 \\ 
        \hline
    \end{tabular}
    \label{table:hyperparams}
    \vspace{0mm}
\end{table}

\begin{figure}
   
    \includegraphics[width=0.85\columnwidth]{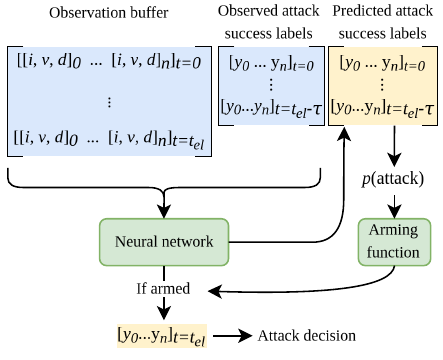}
    \caption{TCML adversary structure (inputs in blue, outputs in yellow). Subscripts $_0..._n$ denote vertex.}
    \vspace{-5mm}
    \label{fig:adversary}
     \vspace{-1mm}
\end{figure}
This neural network is incorporated into the larger adversary model alongside a method of deciding when to ``arm" the adversary and begin to attempt to launch attacks. The full adversary model is illustrated in Figure~\ref{fig:adversary}. 

\subsection{On-line training and operation}
At the start of an attack scenario, the neural network is randomly initialized and trained on-line until it has developed enough confidence in its predictive power to launch an attack. Every timestep within the scenario, the environment is observed and the neural network inputs are calculated as described above. These inputs for every vertex are logged in an observation buffer. Once an amount of time equal to $\tau$ has elapsed following an observation, each set of vertex data can be labeled according to whether an attack at that vertex would have succeeded if launched at the timestep associated with the observation. This allows us to generate labelled training data within the attack scenario, with a delay of $\tau$.

Every timestep, a minibatch of sets of $t_{obs}$ consecutive observations is sampled from the labeled portion of the observation buffer and used to perform a training step of the neural network. The values output from the network are used to update a buffer of outputs associated with each observed input. This output buffer is then used to estimate the probability of the neural network predicting that at least one successful attack is possible in any given timestep (this estimated probability is updated every timestep as the output buffer updates). The estimated attack probability and the time remaining until the time horizon $T$ has elapsed are then used to estimate the probability that the neural network in its current state will launch an attack with time to complete in the remaining time in $T$. Once at least half of $T$ has elapsed and this value drops below a set threshold (0.999 was chosen for our testing), the adversary is ``armed". This process balances the risk of attacking too soon, on the basis of inadequate information, against attacking too late, and running out of time. Once armed, at every timestep the adversary will generate predictions of attack success or failure for each node based on the previous $t_{obs}$ timesteps of observations, and if an attack is predicted to be successful it will launch it. Finally, we record whether the attack succeeds or fails and the scenario ends.

As the ability of the TCML adversary to handle unknown environments and patrol agent strategies was of significant importance, no pre-training was carried out at any point in our testing. The neural network was randomly initialized at the start of every scenario and then trained entirely on-line every time, to demonstrate this ability to generalize to unseen maps and patrol agent strategies.

\section{Simulation testing and results}
\label{sec:sim}

\begin{figure}[H]
   
    \includegraphics[width=\columnwidth]{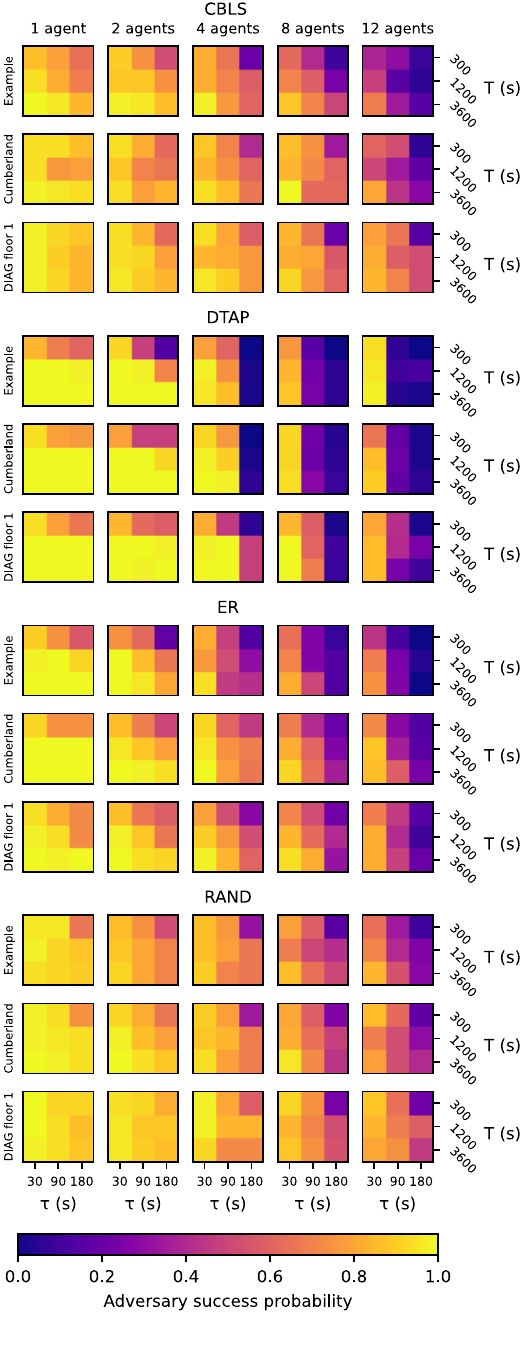}
    \vspace{-0mm}
    \caption{TCML adversary performance against all tested patrol scenarios for varying attack duration $\tau$ and time horizon $T$}
    \label{fig:tcml}
\end{figure}

Our TCML adversary was tested extensively in simulation against the adversary models specified in Section~\ref{sec:adversary_models}. All simulation was carried out in \textit{ROS Patrolling Sim}\footnote{\label{rosnote}\url{http://wiki.ros.org/patrolling\_sim}}~\cite{portugal:patrolsim_paper}, a \textit{Stage}\footnote{\label{stagenote}\url{https://playerproject.github.io/stage/}}-based multi-robot simulator designed for patrolling applications, featuring full LIDAR and odometry simulation in a 2.5D world. To investigate a wide range of patrol team behavior, we simulated three leading decentralized MRP strategies --- DTAP~\cite{farinelli:dta_paper}, CBLS~\cite{portugal:cbls_paper}, and ER~\cite{yan:er_paper} --- alongside a nondeterministic baseline controller (referred to as RAND), in which upon arriving at a vertex of $\mathcal{G}$ each agent randomly selects a neighboring vertex as its next target, while avoiding selecting vertices that other agents have already broadcast their intention to travel to. Our selection of these strategies was motivated by the desire to examine a range of decentralized strategies with fundamentally different design principles, in order to test our TCML adversary against a range of behaviors --- DTAP is based on dynamic auction-based task allocation, CBLS is based on online Bayesian learning, and ER uses a simple mathematical reckoning of the values of possible actions. Additionally, CBLS, DTAP, and ER promise leading levels of idleness-minimization performance, ensuring that our adversary models are tested against highly performance patrol strategies. These strategies were simulated on three different maps provided in \textit{ROS Patrolling Sim} ("Example'', "Cumberland'', and "DIAG floor 1''), with team sizes of 1, 2, 4, 8, and 12 agents. For each combination of strategy, map, and team size, we collected 50 hour-long datasets. These datasets were taken from longer simulation runs to ensure that a wide range of starting agent positions and vertex idlenesses were tested, and to avoid transient startup patrolling behaviors. The datasets were then used to test the five adversary models previously discussed (\textit{random}, \textit{deterministic}, \textit{full-knowledge}, \textit{intelligent (probabilistic)}, and \textit{intelligent (TCML)}), by feeding the pre-recorded patrol data into the adversary models one timestep at a time. Against each dataset, each adversary was tested with time horizons of 300, 1200, and 3600 seconds, and attack durations of 30, 90, and 180 seconds. Adversary successes and failures were recorded, allowing us to estimate success probabilities of the different adversary models with a range of parameters against a range of patrol scenarios. Figure \ref{fig:tcml} shows the observed success probabilities of our TCML adversary against all tested scenarios, and Tables \ref{table:time_horizon}---\ref{table:strategy} show adversary success probabilities for all five adversary models averaged against each scenario variable separately.

\begin{table}[H]
    \setlength\tabcolsep{4.2pt}
    \vspace{-1mm}
    \centering
    % \caption{Average adversary success probabilities for different adversary strategies against time horizon}
    \caption{Success probabilities against time horizon}
    \vspace{-2.5mm}
    \label{table:time_horizon}
    \begin{tabular}{cccccc}
        \hline
        Time horizon (s) & Random & Det. & Full-know. & Int(Pr) & Int(TCML) \\ \hline
        300 & 0.53 & 0.37 & 0.94 & 0.47 & 0.60 \\ 
        1200 & 0.53 & 0.37 & 0.96 & 0.51 & 0.71 \\ 
        3600 & 0.52 & 0.38 & 0.97 & 0.56 & 0.76 \\ 
        \hline
    \end{tabular}
    \vspace{-5mm}
\end{table}

\begin{table}[H]
    \setlength\tabcolsep{4.0pt}
    \vspace{-1mm}
    \centering
    % \caption{Average adversary success probabilities for different adversary strategies against attack duration}
    \caption{Success probabilities against attack duration}
    \vspace{-2.5mm}
    \label{table:attack_duration}
    \begin{tabular}{cccccc}
    \hline
        Attack duration (s) & Random & Det. & Full-know. & Int(Pr) & Int(TCML) \\ \hline
        30 & 0.72 & 0.56 & 1.00 & 0.71 & 0.88 \\ 
        90 & 0.50 & 0.33 & 0.98 & 0.48 & 0.69 \\ 
        180 & 0.35 & 0.23 & 0.90 & 0.35 & 0.49 \\ 
    \hline
    \end{tabular}
    \vspace{-5mm}
\end{table}

\begin{table}[H]
    \setlength\tabcolsep{4.2pt}
    \vspace{-1mm}
    \centering
    % \caption{Average adversary success probabilities for different adversary models against number of patrol agents}
    \caption{Success probabilities against number of patrol agents}
    \vspace{-2.5mm}
    \label{table:n_agents}
    \begin{tabular}{cccccc}
    \hline
        No. agents & Random & Det. & Full-know. & Int(Pr) & Int(TCML) \\ \hline
        1 & 0.82 & 0.62 & 1.00 & 0.81 & 0.91 \\ 
        2 & 0.69 & 0.53 & 1.00 & 0.71 & 0.83 \\ 
        4 & 0.53 & 0.36 & 0.97 & 0.49 & 0.70 \\ 
        8 & 0.35 & 0.21 & 0.93 & 0.31 & 0.55 \\ 
        12 & 0.25 & 0.17 & 0.89 & 0.26 & 0.44 \\ 
    \hline
    \end{tabular}
    \vspace{-5mm}
\end{table}

\begin{table}[H]
    \setlength\tabcolsep{4.2pt}
    \vspace{-1mm}
    \centering
    % \caption{Average adversary success probabilities for different adversary strategies against map}
    \caption{Success probabilities against map}
    \vspace{-2.5mm}
    \label{table:map}
    \begin{tabular}{cccccc}
    \hline
        Map & Random & Det. & Full-know. & Int(Pr) & Int(TCML) \\ \hline
        ``Example" & 0.44 & 0.38 & 0.93 & 0.45 & 0.62 \\ 
        ``Cumberland" & 0.54 & 0.39 & 0.96 & 0.53 & 0.70 \\ 
        ``DIAG floor 1" & 0.59 & 0.36 & 0.98 & 0.57 & 0.74 \\ 
    \hline
    \end{tabular}
    \vspace{-5mm}
\end{table}

\begin{table}[H]
    \setlength\tabcolsep{4.2pt}
    \vspace{-1mm}
    \centering
      \caption{Success probabilities against patrol strategy}
    \vspace{-2.5mm}
    \label{table:strategy}
    \begin{tabular}{cccccc}
    \hline
        Patrol strategy & Random & Det. & Full-know. & Int(Pr) & Int(TCML) \\ \hline
        DTAP & 0.42 & 0.57 & 0.86 & 0.41 & 0.64 \\ 
        ER & 0.52 & 0.42 & 0.99 & 0.50 & 0.66 \\ 
        CBLS & 0.57 & 0.25 & 0.99 & 0.58 & 0.71 \\ 
        RAND & 0.60 & 0.25 & 1.00 & 0.58 & 0.74 \\ 
    \hline
    \end{tabular}
 
\end{table}

\newpage

\section{Real-world testing and results}

\begin{figure}[H]
    \centering
    \includegraphics[width=0.95\linewidth]{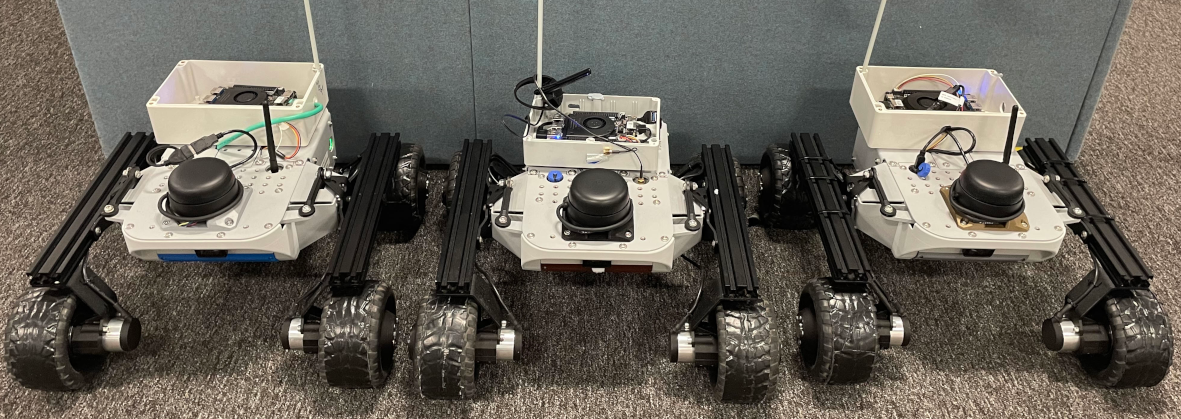}
    \caption{The three Leo Rover robots used for testing}
    \label{fig:rovers}
\end{figure}

\begin{figure}[H]
    \centering
    \includegraphics[width=0.95\linewidth]{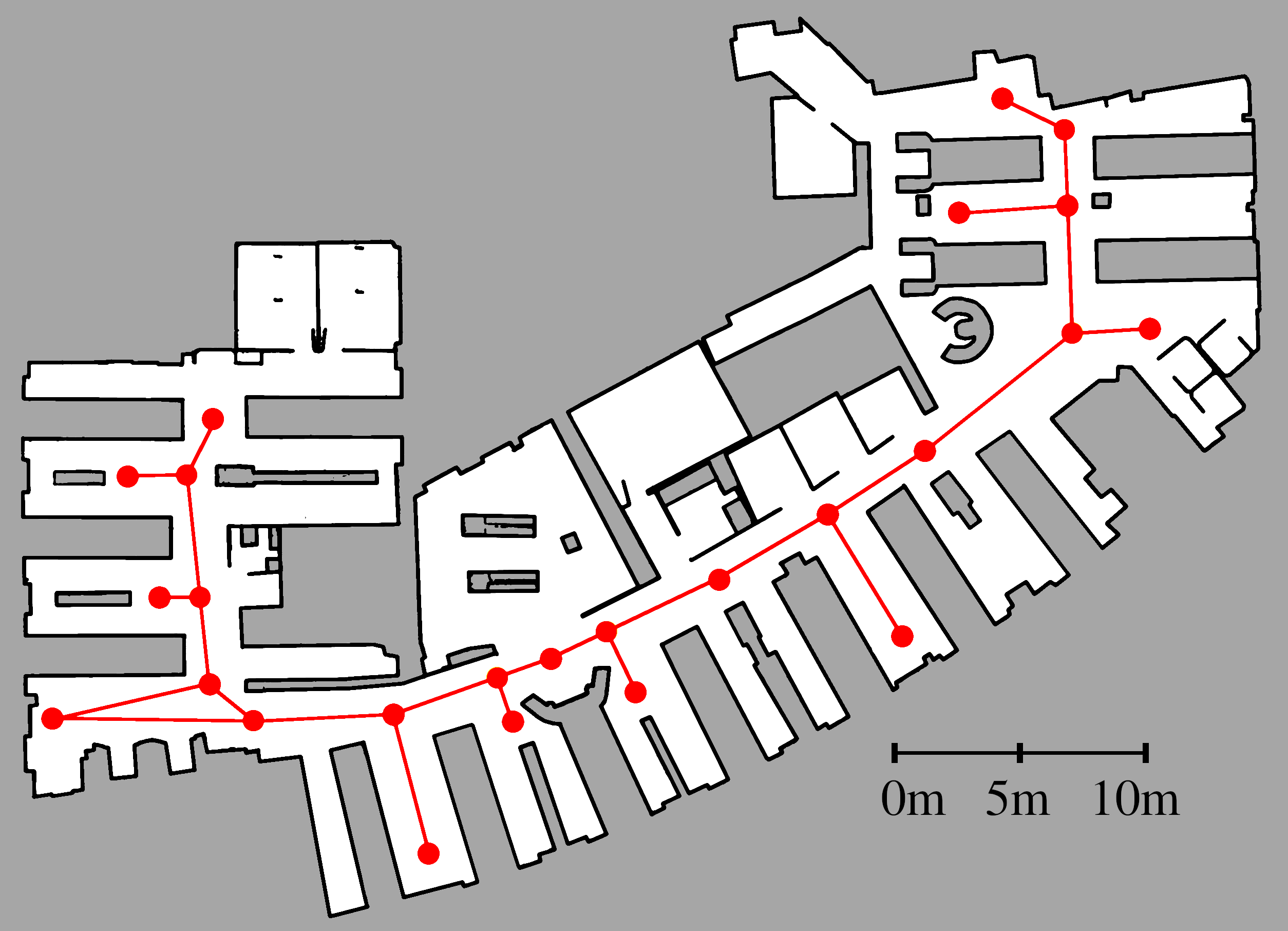}
    \caption{LIDAR-generated office floorplan overlaid with patrol graph}
    \label{fig:office_map}
\end{figure}

To further validate our adversary model, we carried out testing on real-world patrol data. Three LIDAR-equipped Leo Rover\footnote{\label{leonote}\url{https://www.leorover.tech/}} robots, shown in Figure~\ref{fig:rovers}, were programmed with the nondeterministic ``RAND" patrol strategy (described in Section~\ref{sec:sim}) to autonomously patrol an office environment. The LIDAR-generated map of the environment, overlaid with the patrol graph used, is shown in Figure~\ref{fig:office_map}. A total of 25 patrol data windows of length 1200 s were recorded --- as with our simulation tests, these were taken from longer continuous runs to avoid transient startup behaviors and ensure a wide range of initial agent positions and vertex idlenesses. Due to practical limits on battery life, we were unable to record enough sufficiently long runs to allow for testing on 3600 s windows.

These data were then passed to our adversary models to assess their performance against a real-world patrol team. The resultant adversary success rates are shown in Table~\ref{table:real_results}. It is worth noting that the deployment of the adversary model functioned the same in these tests as in the tests on simulated data, and the difference is the source of the patrol data being examined --- as discussed in Section~\ref{sec:discussion}, we do not yet consider the physical real-time deployment of an intelligent adversary.

\begin{table}[]
    \centering
    \caption{Adversary success probabilities against real patrol team}
    \vspace{-2mm}
    \label{table:real_results}
    \setlength\tabcolsep{3pt}
    \begin{tabular}{cccccc}
        \multicolumn{6}{c}{$T$=300s}  \\
        \hline
        Attack duration (s) & Random & Det. & Full-know. & Int(Pr) & Int(TCML) \\ 
        \hline
        30 & 0.84 & 0.92 & 1.00 & 0.80 & 0.92 \\ 
        90 & 0.68 & 0.40 & 1.00 & 0.52 & 0.76 \\ 
        180 & 0.36 & 0.08 & 1.00 & 0.36 & 0.36 \\ 
        \hline
        \multicolumn{6}{c}{$T$=1200s}  \\
        \hline
         Attack duration (s) & Random & Det. & Full-know. & Int(Pr) & Int(TCML) \\ 
        \hline
        30 & 0.80 & 1.00 & 1.00 & 0.92 & 0.92 \\ 
        90 & 0.72 & 0.44 & 1.00 & 0.56 & 0.84 \\ 
        180 & 0.56 & 0.24 & 1.00 & 0.28 & 0.68 \\ 
        \hline
    \end{tabular}
    \vspace{-6mm}
\end{table}

\section{Discussion}

\label{sec:discussion}

Here we have demonstrated the value of a time-constrained machine-learning based system for in-simulation vulnerability analysis of an automated system, specifically a multi-robot patrol system. Our results show that our ML-based intelligent adversary is a significant improvement over the three realistic (i.e. not full-knowledge) adversaries tested in time-constrained scenarios, and thus presents a more stringent test for assessing patrol strategy performance in adversarial scenarios. Additionally, while our pre-existing probabilistic adversary model has previously been shown to offer significant improvements over the random and deterministic models with six hours of learning time~\cite{ward:benchmarking_method}, in the shorter time horizons examined here it was not found to offer significant improvement over the random adversary. The performance of the deterministic adversary was notably poor against patrol strategies that frequently caused agents to immediately return to a vertex having just left it, while its performance against DTAP (which does not exhibit this behavior) is better (Table~\ref{table:strategy}).

Our results suggest that, despite the increased complexity of the TCML adversary compared to the probabilistic model, it is sufficiently sample-efficient to outperform realistic baselines for even the shortest time horizons tested. The neural network was designed with this as the goal, with a relatively light-weight architecture and more aggressive regularization than might be typically desirable. We do not doubt that a more complex or differently tuned network might be able to achieve superior performance for longer time horizons, but as the goal of this work was to develop an adversary model that represented a plausible real-world threat, having it be highly sample-efficient and able to perform well in shorter time horizons was our priority. The time horizons chosen for testing reflect this, as during initial testing we found that the performance of the TCML adversary does not improve significantly past 3600 s, as a result of our prioritization of performance for shorter time horizons.

DTAP was found to typically perform the best against adversaries (lowest adversary success probability), with the notable exception of performance against the deterministic adversary, as mentioned above. The relatively poor performances of the random and full-knowledge adversaries against DTAP suggest that it offers fewer potential attack windows than other strategies. The RAND patrol strategy, despite being non-deterministic, showed the worst performance (highest adversary success probability) against the probabilistic and TCML adversaries of all strategies examined. While the decisions of patrol agents in a non-deterministic strategy are in general not possible to predict and therefore not possible for an adversary to learn to anticipate, the possibility for patrol agents to leave large areas of the patrol graphs unvisited for long periods of time with no nearby agents (unlike the other strategies considered, which attempt to visit all vertices frequently and disperse agents evenly) was clearly easy to exploit by both of the learning adversary models. This suggests that any future efforts to develop patrol strategies specifically designed to defeat intelligent adversaries must give at least some weighting to efficient coverage behavior, even at the expense of unpredictability.

Our tests against real-world patrol data, while not as extensive as our tests against simulated data, support these conclusions. Due to the relatively small number of patrol datasets we were able to capture, drawing strong conclusions would be inappropriate, but the same general trends can be observed as were present for our simulated data --- our TCML adversary model outperforming the other models (except the theoretical full-knowledge adversary), with performance increasing with the time horizon. Due to the degree of completeness of \textit{ROS Patrolling Sim}, we do not anticipate that different trends will emerge for real-world data.

One limitation of the adversary models considered in this work is that they consider all vertices equally for attack. In a real-world scenario, it may be desirable for an attacker to only target certain areas of the environment, or for required attack durations to vary depending on the target vertex, or any number of scenario-specific considerations. Such modifications would be straightforward to apply to our TCML adversary model, but because they would be so scenario-specific we consider them to be outside the scope of this work. A second limitation is that the adversary models consider not launching an attack within the time horizon to be equivalent to launching an attack which fails. Again, the relative negative utility of an attack failing versus no attack being launched would be very straightforward to modify in our model but also very scenario specific, and so we have not considered it. The remaining major obstacles to this model being deployed in real-time in the real world --- i.e. the ability to inconspicuously observe a patrol system in order to gather the requisite input data, and the ability to physically launch attacks against the patrol system --- are not considered in the current work, since they would again be very scenario specific. Nevertheless, this work is a meaningful step toward more realistic performance analysis and future design of patrolling strategies, as despite real-world limitations we have demonstrated that this can be carried out effectively in simulation.

These results support the utility of this approach as a simulated red-team approach for a robotic system. This is no substitute for a true red-teaming of an integrated security system, which would include potential exploitation of human elements, cybersecurity, and physical security after a lengthy period of planning and observation, but this compressed simulation approach can serve to highlight potential vulnerabilities in automated aspects of a larger system.

\section{Conclusions}

We present a new Time-Constrained ML-based (TCML) adversary model for the Multi-Robot Patrolling problem, designed to function with limited observation time before an attack must be made. Following comparison to existing models both in simulation and against real-world patrol data, our adversary model is shown to significantly outperform realistic baselines in time-limited scenarios. As such, this model can act as a more robust and plausible test of multi-robot patrol strategy performance when encountering adversaries in real-world conditions, and may offer insight into design considerations for future patrol strategy development. This simulated red-teaming affords us insight into multi-robot patrol performance that would not be available from other forms of analysis, demonstrating the potential utility of such approaches in broader automation security contexts.

\bibliographystyle{IEEEtran}
\bibliography{IEEEabrv,main}

\begin{thebibliography}{10}
\providecommand{\url}[1]{#1}
\csname url@samestyle\endcsname
\providecommand{\newblock}{\relax}
\providecommand{\bibinfo}[2]{#2}
\providecommand{\BIBentrySTDinterwordspacing}{\spaceskip=0pt\relax}
\providecommand{\BIBentryALTinterwordstretchfactor}{4}
\providecommand{\BIBentryALTinterwordspacing}{\spaceskip=\fontdimen2\font plus
\BIBentryALTinterwordstretchfactor\fontdimen3\font minus \fontdimen4\font\relax}
\providecommand{\BIBforeignlanguage}[2]{{%
\expandafter\ifx\csname l@#1\endcsname\relax
\typeout{** WARNING: IEEEtran.bst: No hyphenation pattern has been}%
\typeout{** loaded for the language `#1'. Using the pattern for}%
\typeout{** the default language instead.}%
\else
\language=\csname l@#1\endcsname
\fi
#2}}
\providecommand{\BIBdecl}{\relax}
\BIBdecl

\bibitem{review_added:1}
F.~Martín, E.~Soriano-Salvador, J.~M. Guerrero, G.~{Guardiola Múzquiz}, J.~C. Manzanares, and F.~J. Rodríguez, ``Towards a robotic intrusion prevention system: Combining security and safety in cognitive social robots,'' \emph{Robotics and Autonomous Systems}, vol. 190, p. 104959, 2025.

\bibitem{review_added:2}
D.~Nurchalifah, S.~Blumenthal, L.~Lo~Iacono, and N.~Hochgeschwender, ``Analysing the safety and security of a uv-c disinfection robot,'' in \emph{2023 IEEE International Conference on Robotics and Automation (ICRA)}, 2023, pp. 12\,729--12\,736.

\bibitem{karnik:foundation_models}
\BIBentryALTinterwordspacing
S.~Karnik, Z.-W. Hong, N.~Abhangi, Y.-C. Lin, T.-H. Wang, C.~Dupuy, R.~Gupta, and P.~Agrawal, ``Embodied red teaming for auditing robotic foundation models,'' 2025. [Online]. Available: \url{https://arxiv.org/abs/2411.18676}
\BIBentrySTDinterwordspacing

\bibitem{abhangi:redteam}
N.~Abhangi, ``Red teaming language conditioned robotic behavior,'' Doctoral Dissertation, Massachusetts Institute of Technology, 2024.

\bibitem{majumdar:redteam}
\BIBentryALTinterwordspacing
A.~Majumdar, M.~Sharma, D.~Kalashnikov, S.~Singh, P.~Sermanet, and V.~Sindhwani, ``Predictive red teaming: Breaking policies without breaking robots,'' 2025. [Online]. Available: \url{https://arxiv.org/abs/2502.06575}
\BIBentrySTDinterwordspacing

\bibitem{mansfield:redteam}
S.~Mansfield-Devine, ``The best form of defence – the benefits of red teaming,'' \emph{Computer Fraud \& Security}, vol. 2018, no.~10, pp. 8--12, 2018.

\bibitem{machade:idleness_paper}
A.~Machado, G.~Ramalho, J.~Zucker, and A.~Drogoul, ``Multi-agent patrolling: An empirical analysis of alternative architectures,'' in \emph{Proceedings of the 3rd International Conference on Multi-Agent Based Simulation II}, Bologna, Italy, July 2002, pp. 155--170.

\bibitem{chevaleyre:tsp_paper}
Y.~Chevaleyre, ``Theoretical analysis of the multi-agent patrolling problem,'' in \emph{Proceedings of the IEEE/WIC/ACM International Conference on Intelligent Agent Technology}, September 2004, pp. 302--308.

\bibitem{Afshani2020}
P.~Afshani, M.~de~Berg, K.~Buchin, J.~Gao, M.~Löffler, A.~Nayyeri, B.~Raichel, R.~Sarkar, H.~Wang, and H.-T. Yang, ``Approximation algorithms for multi-robot patrol-scheduling with min-max latency,'' \emph{14th International Workshop on the Algorithmic Foundations of Robotics}, 2020.

\bibitem{Afshani2022}
------, ``On cyclic solutions to the min-max latency multi-robot patrolling problem,'' \emph{38th International Symposium on Computational Geometry}, 2022.

\bibitem{Rosenkrantz:tsp}
D.~J. Rosenkrantz, R.~E. Stearns, and P.~M. Lewis, ``Approximate algorithms for the traveling salesperson problem,'' in \emph{15th Annual Symposium on Switching and Automata Theory (swat 1974)}, 1974, pp. 33--42.

\bibitem{farinelli:dta_paper}
A.~Farinelli, L.~Iocchi, and D.~Nardi, ``Distributed on-line dynamic task assignment for multi-robot patrolling,'' \emph{Autonomous Robots}, vol.~41, no.~6, pp. 1321--1345, August 2017.

\bibitem{portugal:sebs_paper}
D.~Portugal and R.~P. Rocha, ``Distributed multi-robot patrol: A scalable and fault-tolerant framework,'' \emph{Robotics and Autonomous Systems}, vol.~61, no.~12, pp. 1572--1587, December 2013.

\bibitem{portugal:cbls_paper}
D.~Portugal and R.~P. Rocha\vspace{0mm}, ``Cooperative multi-robot patrol with bayesian learning,'' \emph{Autonomous Robots}, vol.~40, no.~5, pp. 929--953, June 2016.

\bibitem{yan:er_paper}
C.~Yan and T.~Zhang, ``Multi-robot patrol: A distributed algorithm based on expected idleness,'' \emph{International Journal of Advanced Robotic Systems}, vol.~13, no.~6, November 2016.

\bibitem{li:gnn}
Q.~Li, F.~Gama, A.~Ribeiro, and A.~Prorok, ``Graph neural networks for decentralized multi-robot path planning,'' in \emph{Proceedings of the 2020 International Conference on Intelligent Robots and Systems (IROS)}, 10 2020, pp. 11\,785--11\,792.

\bibitem{zhou:gnn}
Y.~Zhou, J.~Xiao, Y.~Zhou, and G.~Loianno, ``Multi-robot collaborative perception with graph neural networks,'' \emph{IEEE Robotics and Automation Letters}, vol.~PP, pp. 1--1, 01 2022.

\bibitem{tolstaya:gnn}
E.~Tolstaya, F.~Gama, J.~Paulos, G.~Pappas, V.~Kumar, and A.~Ribeiro, ``Learning decentralized controllers for robot swarms with graph neural networks,'' in \emph{Proceedings of the Conference on Robot Learning}, ser. Proceedings of Machine Learning Research, L.~P. Kaelbling, D.~Kragic, and K.~Sugiura, Eds., vol. 100.\hskip 1em plus 0.5em minus 0.4em\relax PMLR, 30 Oct--01 Nov 2020, pp. 671--682.

\bibitem{tolstaya:gnn_patrol}
E.~V. Tolstaya, J.~Paulos, V.~R. Kumar, and A.~Ribeiro, ``Multi-robot coverage and exploration using spatial graph neural networks,'' \emph{2021 IEEE/RSJ International Conference on Intelligent Robots and Systems (IROS)}, pp. 8944--8950, 2020.

\bibitem{ward:suns_mns}
\BIBentryALTinterwordspacing
J.~C. Ward, R.~McConville, and E.~R. Hunt, ``Lightweight decentralized neural network-based strategies for multi-robot patrolling,'' in \emph{Proceedings of the 40thth ACM/SIGAPP Symposium on Applied Computing (SAC '25)}, April 2025, in press. [Online]. Available: \url{https://arxiv.org/abs/2412.11916}
\BIBentrySTDinterwordspacing

\bibitem{guo:marl_approach}
L.~Guo, H.~Pan, X.~Duan, and J.~He, ``Balancing efficiency and unpredictability in multi-robot patrolling: A marl-based approach,'' in \emph{2023 IEEE International Conference on Robotics and Automation (ICRA)}, 2023, pp. 3504--3509.

\bibitem{zhou:learning_patrol}
X.~Zhou, W.~Wang, T.~Wang, Y.~Lei, and F.~Zhong, ``Bayesian reinforcement learning for multi-robot decentralized patrolling in uncertain environments,'' \emph{IEEE Transactions on Vehicular Technology}, vol.~68, no.~12, pp. 11\,691--11\,703, 2019.

\bibitem{Hsu2014}
S.-B. Hsu, C.-C. Lien, C.-H. Lee, C.-C. Han, S.-P. Chen, and Y.-L. Chang, ``Robot patrolling using sensor data and image features,'' in \emph{2014 International Conference on Information Science, Electronics and Electrical Engineering}, vol.~3, 2014, pp. 1737--1741.

\bibitem{Ghosh2021}
A.~Ghosh, A.~Dutta, and B.~Sotolongo, ``Minimalist coverage and energy-aware tour planning for a mobile robot,'' in \emph{2022 IEEE 18th International Conference on Automation Science and Engineering (CASE)}, 2022, pp. 2056--2061.

\bibitem{agmon:full_knowledge}
N.~Agmon and S.~Kaminka, G. A.~Kraus, ``Multi-robot adversarial patrolling: Facing a full-knowledge opponent,'' \emph{Journal of Artificial Intelligence Research}, vol.~42, no.~1, pp. 887--916, September 2011.

\bibitem{duan:markov}
X.~Duan and F.~Bullo, ``Markov chain-based stochastic strategies for robotic surveillance,'' \emph{Annual Review of Control, Robotics, and Autonomous Systems}, vol.~4, pp. 243--264, May 2021.

\bibitem{alam:markov}
T.~Alam, ``Decentralized and nondeterministic multi-robot area patrolling in adversarial environments,'' \emph{International Journal of Computer Applications}, vol. 156, no.~2, pp. 1--8, December 2016.

\bibitem{alam:markov2}
T.~Alam, M.~M. Rahman, P.~Carrillo, L.~Bobadilla, and B.~Rapp, ``Stochastic multi-robot patrolling with limited visibility,'' \emph{Journal of Intelligent \& Robotic Systems}, vol.~97, pp. 411--429, 2020.

\bibitem{basiloco:markov}
N.~Basilico and S.~Carpin, ``Balancing unpredictability and coverage in adversarial patrolling settings,'' in \emph{Algorithmic Foundations of Robotics XIII}, M.~Morales, L.~Tapia, G.~S{\'a}nchez-Ante, and S.~Hutchinson, Eds.\hskip 1em plus 0.5em minus 0.4em\relax Springer International Publishing, 2020, pp. 762--777.

\bibitem{yang:gametheory}
H.-T. Yang, S.-Y. Tsai, K.~S. Liu, S.~Lin, and J.~Gao, ``Patrol scheduling against adversaries with varying attack durations,'' in \emph{Proceedings of the 18th International Conference on Autonomous Agents and Multiagent Systems (AAMAS 2019)}, Montreal, Canada, May 2019.

\bibitem{asghar:markov2}
A.~B. Asghar and S.~S. L., ``A patrolling game for adversaries with limited observation time,'' in \emph{Proceedings of the 2018 Conference on Decision and Control (CDC)}, Miami Beach, USA, December 2018, pp. 3305--3310.

\bibitem{parachuri:stackelberg_games}
P.~Paruchuri, J.~P. Pearce, S.~Kraus, and J.~Marecki, ``Playing games for security: An efﬁcient exact algorithm for solving bayesian stackelberg games,'' in \emph{Proceedings of the 7th International Joint Conference on Autonomous Agents and Multiagent Systems}, Estoril, Portugal, May 2008, pp. 895--902.

\bibitem{clempner:game_theory}
J.~B. Clempner, ``A continuous-time markov stackelberg security game approach for reasoning about real patrol strategies,'' \emph{International Journal of Control}, vol.~91, no.~11, pp. 2494--2510, 2018.

\bibitem{Buermann:fence_patrol}
J.~Buermann and J.~Zhang, ``Multi-robot adversarial patrolling strategies via lattice paths,'' \emph{Artificial Intelligence}, vol. 311, 2022.

\bibitem{agmon:fence_patrol}
N.~Agmon and S.~Kaminka, G. A.~Kraus, ``Multi-robot adversarial patrolling: Facing a full-knowledge opponent,'' \emph{Journal of Artificial Intelligence Research}, vol.~42, no.~1, pp. 887--916, September 2011.

\bibitem{sless:coordinated_attacks}
E.~Sless, N.~Agmon, and S.~Kraus, ``Multi-robot adversarial patrolling: Facing coordinated attacks,'' in \emph{Proceedings of the 2014 International Conference on Autonomous Agents and Multi-Agent Systems}, Paris, France, May 2014, pp. 2093--1100.

\bibitem{sless:sequential_attacks}
E.~S. Lin, N.~Agmon, and S.~Kraus, ``Multi-robot adversarial patrolling: Handling sequential attacks,'' \emph{Artificial Intelligence}, vol. 274, February 2019.

\bibitem{asghar:markov}
A.~B. Asghar and S.~S. L., ``Stochastic patrolling in adversarial settings,'' in \emph{Proceedings of the 2016 American Control Conference (ACC)}, Boston, USA, July 2019, pp. 6435--6440.

\bibitem{ward:benchmarking_method}
J.~C. Ward and E.~R. Hunt, ``An empirical method for benchmarking multi-robot patrol strategies in adversarial environments,'' in \emph{Proceedings of the 38th ACM/SIGAPP Symposium on Applied Computing}, Tallin, Estonia, March 2023, pp. 787--790.

\bibitem{portugal:patrolsim_paper}
D.~Portugal, L.~Iocchi, and A.~Farinelli, ``A {ROS}-based framework for simulation and benchmarking of multi-robot patrolling algorithms,'' in \emph{Robot Operating System (ROS)}.\hskip 1em plus 0.5em minus 0.4em\relax New {Y}ork: Springer, 2018, vol.~3, pp. 3--28.

\end{thebibliography}

\end{document}